\documentclass[10pt,twocolumn,letterpaper]{article}

\PassOptionsToPackage{dvipsnames,svgnames,table,xcdraw}{xcolor}
\usepackage[pagenumbers]{cvpr}

\usepackage{times}
\usepackage{epsfig}
\usepackage{graphicx}
\usepackage{tabularx}
\usepackage{threeparttable}

\providecommand{\na}{\textemdash}
\usepackage{amsmath}
\usepackage{amssymb}
\usepackage{multirow}
\usepackage{bbold}
\usepackage{array}
\usepackage{booktabs}
\usepackage{algorithm}
\usepackage{algpseudocode}
\usepackage{bbding}
\usepackage{makecell}
\usepackage{balance}  
\usepackage{diagbox}
\usepackage{colortbl}
\newcommand{\PreserveBackslash}[1]{\let\temp=\\#1\let\\=\temp}
\newcolumntype{C}[1]{>{\PreserveBackslash\centering}p{#1}}
\newcolumntype{R}[1]{>{\PreserveBackslash\raggedleft}p{#1}}
\newcolumntype{L}[1]{>{\PreserveBackslash\raggedright}p{#1}}

\newcommand{\secref}[1]{Section~\ref{sec:#1}}
\newcommand{\figref}[1]{Figure~\ref{fig:#1}}

\newcommand{\Paragraph}[1]{\noindent\textbf{#1}}

\definecolor{confgray}{gray}{0.65}      % 灰色 (之前的)
\definecolor{metricred}{rgb}{0.85, 0, 0}  % 定义一个醒目的红色 (R,G,B)
\newlength\secmargin
\newlength\subsecmargin
\newlength\paramargin
\newlength\figmargin
\newlength\eqmargin
\definecolor{cvprblue}{rgb}{0.21,0.49,0.74}
\usepackage[pagebackref,breaklinks,colorlinks,allcolors=cvprblue]{hyperref}

\def\confName{CVPR}
\def\confYear{2027}

\title{NavDecision: Parallel Multi-Step Action Prediction with Persistent Self-Referential Learning for Vision-and-Language Navigation}

\author{
  Kai Wu$^{1}$ \quad Yufan Hu$^{2}$ \quad Quan Fang$^{3}$ \quad
  Boning Zhang$^{1}$ \quad Junyu Gao$^{1}$\thanks{Corresponding author.}\\[4pt]
  $^{1}$Institute of Automation, Chinese Academy of Sciences\\
  $^{2}$School of Intelligence Science and Technology,
  University of Science and Technology Beijing\\
  $^{3}$School of Artificial Intelligence,
  Beijing University of Posts and Telecommunications\\[4pt]
  {\small\ttfamily wukai2025@ia.ac.cn \quad huyufanqaixuan@gmail.com \quad
  qfang@bupt.edu.cn}\\
  {\small\ttfamily zhangboning2025@ia.ac.cn \quad junyu.gao@nlpr.ia.ac.cn}
}

\begin{document}
\maketitle
\begin{abstract}
Vision-and-language navigation (VLN) requires timely decisions based on instructions and visual observations. However, autoregressive VLN models generate action descriptions token by token, introducing substantial inference overhead. Decision models offer an efficient alternative but face challenges in maintaining short-horizon action coherence and long-horizon goal alignment. To address these challenges, we propose NavDecision, an efficient multi-step VLN decision model. We introduce parallel multi-step action prediction (PMAP) to predict multiple action distributions from a shared vision-language state, and memory-augmented self-referential policy optimization (M-SRPO) to learn from unsuccessful trajectories using successful trajectories as references. A cross-iteration success memory further accumulates historical successful trajectories to provide informative training feedback. NavDecision achieves success rates of 64.60\% on R2R-CE and 64.57\% on RxR-CE with a 4B backbone. Compared with existing autoregressive VLN methods, it delivers significant inference speedups of up to $22.28\times$ and $18.96\times$ while maintaining comparable navigation performance.
\end{abstract}    
\section{Introduction} \label{sec:intro}   
% Introduction body fragment. The parent document supplies the top-level section.
% % Introduction motivation figure (single column).
% % Place this file and fig1(1).pdf in the project's fig/ directory.
% % Insert with \input{fig/fig1}; cite with \figref{motivation}.
% \begin{figure}[t]
%     \centering
%     \includegraphics[width=\columnwidth]{fig/pdf/fig1.pdf}
%     \caption{\textbf{From long-horizon experience to fast local decisions.} (a) Autoregressive VLN decodes action text sequentially; terminal binary success rewards give failures identical feedback regardless of progress. (b) Our method predicts multiple action distributions in parallel and learns from failures using successes retained across training iterations, incorporating long-horizon experience into the policy.}
%     \label{fig:motivation}
% \end{figure}

% Introduction motivation figure (double-column).
% Place this file and fig1.pdf in the project's fig/ directory.
% Insert with \input{fig/fig1}; cite with \figref{motivation}.
\begin{figure}[t]
    \centering
    \includegraphics[width=\columnwidth]{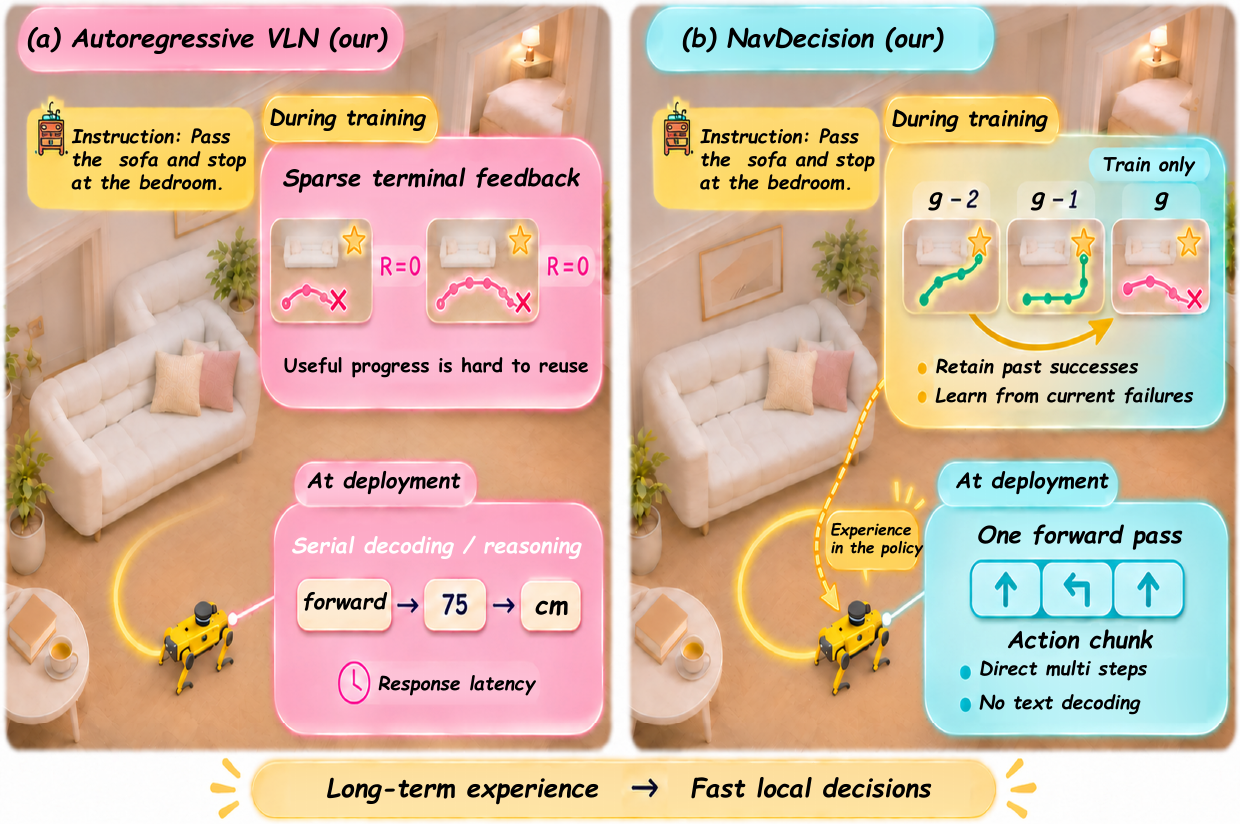}
    \caption{\textbf{Autoregressive VLN vs. NavDecision.}
(a) Autoregressive VLN generates action text sequentially, while terminal binary rewards provide identical feedback for unsuccessful trajectories.
(b) NavDecision predicts multiple action distributions in parallel and learns from historical successes across training iterations.}
    \label{fig:motivation}
\end{figure}

Vision-and-language navigation (VLN) requires agents to reach goals using natural language instructions and visual observations~\cite{anderson2018vln,krantz2020vlnce}. Autoregressive vision-language models (VLMs) use pretrained cross-modal representations to ground instructions in visual scenes, improving navigation in unseen environments and sim-to-real transfer~\cite{zhang2024navid,cheng2025navila}. However, as shown in \figref{motivation}(a), autoregressive VLN models decode action descriptions such as ``move forward 75~cm'' token by token before converting them into executable actions~\cite{cheng2025navila}. Action latency thus depends on output length, while navigation requires timely responses to observations~\cite{sheng2026navjev}. In contrast, decision models directly predict distributions over candidate options~\cite{almeida2026jev,li2026llmasjev}, enabling low-latency execution in time-sensitive interactive tasks~\cite{zhang2026jevmobile}. We explore direct action prediction for VLN to reduce generation overhead while preserving vision-language understanding. However, directly applying decision models to VLN requires addressing both the temporal coherence of consecutive actions and long-horizon goal alignment.

Although decision models reduce the overhead of autoregressive generation, designing an effective decision model for VLN remains challenging. First, single-step primitive-action selection avoids token-by-token decoding but requires one model call per action and does not explicitly model temporal dependencies across actions, making it difficult to maintain action coherence. Autoregressive multi-step policies generate several actions sequentially in one decoding run, conditioning on preceding outputs to capture temporal dependencies~\cite{zhang2026activevln,hong2026simplevln}. Alternatively, decision models can treat predefined fixed action chunks as candidate options, but this restricts action combinations to a fixed library, whose expansion increases the candidate space and decision complexity. Waypoint-based decision methods, in turn, depend on the quality and coverage of externally predicted candidates~\cite{sheng2026navjev}. We therefore seek to predict multi-step action distributions from a shared vision-language state, capturing short-horizon temporal dependencies for coherent execution at low inference cost.

Second, agents must track task progress under changing observations to keep local actions aligned with long-horizon goals. Autoregressive policies can scale test-time computation through explicit reasoning or search to improve navigation success~\cite{lin2025navcot,fried2018speaker}. To retain low inference overhead, long-horizon exploration experience needs to be internalized in the local policy during training. Online reinforcement learning optimizes policies using task-level rewards from complete navigation trajectories, including terminal binary success rewards~\cite{zhang2026activevln,hong2026simplevln}. However, decision models lack explicit test-time computation scaling, making it difficult to consistently discover successful trajectories during long-horizon exploration. Consequently, successful trajectories may be scarce, and terminal binary success rewards provide identical feedback for unsuccessful trajectories despite differences in their progress toward the goal. This limitation highlights the need for more informative trajectory-level feedback to learn from unsuccessful trajectories and improve local action policies.

We propose NavDecision, an efficient multi-step VLN decision model trained with cross-iteration self-referential reinforcement learning. As shown in \figref{motivation}(b), we predict multiple action distributions in parallel from a shared vision-language state. Temporally causal interactions among prediction features and joint multi-step supervision capture short-horizon temporal dependencies. Inspired by self-referential policy optimization (SRPO)~\cite{fei2026srpo}, we encode complete navigation trajectories with a frozen world model. Using environment-verified successful trajectories as references, we combine bounded similarity-based rewards for unsuccessful trajectories with terminal binary success rewards for online reinforcement learning. A cross-iteration success memory combines current-batch and historical successful trajectories, providing references when the current batch contains no successful trajectories and incorporating new successful trajectories as training proceeds. This internalizes long-horizon exploration experience during training while retaining fast action prediction at deployment.
Our main contributions are summarized as follows:
\begin{itemize} 
    \item We extend single-step option selection to parallel multi-step action prediction. Using a shared vision-language state representation, we learn short-horizon temporal dependencies through temporally causal interactions among prediction features and joint multi-step supervision. A single forward pass outputs action distributions for multiple steps, supporting coherent navigation actions and efficient decision-making.

    \item We design a reinforcement learning reward for decision models that uses current-batch and historical successful trajectories to provide more informative training feedback for unsuccessful trajectories. This internalizes long-horizon exploration experience in the decision policy. The world model and cross-iteration success memory are used only during training and do not participate in action inference at deployment.

    \item
    NavDecision achieves 64.60\% and 64.57\% SR on R2R-CE and RxR-CE with a 4B backbone. It features adjustable inference efficiency, enabling up to $22.28\times$ and $18.96\times$ speedups over autoregressive VLN baselines while maintaining competitive navigation performance.
\end{itemize}

\section{Related Work}\label{sec:related_work}
% Related-work body fragment. The parent document supplies the top-level section.

\Paragraph{Vision-and-Language Navigation.}
Vision-and-language navigation has progressed from R2R and RxR benchmarks to low-level action execution in continuous environments~\cite{anderson2018vln,ku2020rxr,krantz2020vlnce}.
Prior work improves generalization through instruction synthesis, environment perturbations, and large-scale data augmentation~\cite{fried2018speaker,tan2019envdrop,wang2023scalevln}.
Cross-modal pretraining and history encoding strengthen state representations~\cite{hao2020prevalent,hong2021vlnbert,chen2021hamt}; topological planning supports global path reasoning~\cite{chen2022duet}, and waypoint prediction connects planning with local continuous control~\cite{hong2022bridging,an2025etpnav}.
Navigation methods based on large models explore explicit
reasoning and map-based planning~\cite{zhou2024navgpt,chen2024mapgpt,zhou2025navgpt2,lin2025navcot},
as well as video-to-action modeling and hierarchical
control~\cite{zhang2024navid,zhang2025uninavid,cheng2025navila,wei2026dualvln}.
Recent methods use action chunking to predict actions across multiple time steps~\cite{zhu2026navida,hong2026simplevln}, while dataset aggregation~\cite{ross2011dagger}, reinforcement fine-tuning with expert-action supervision~\cite{qi2025vlnr1}, and multi-turn online reinforcement learning~\cite{zhang2026activevln,hong2026simplevln} improve policies.
Sequential autoregressive decoding introduces additional action latency.
Although NavJev directly selects candidate waypoints to avoid autoregressive action generation, it relies on an external waypoint predictor~\cite{sheng2026navjev}.
In contrast, we introduce parallel multi-step action prediction over discrete primitive actions, eliminating sequential decoding and reducing model calls.

\Paragraph{Decision Models.}
Decision models predict probability distributions over candidate options, enabling structured decisions without autoregressive generation of action descriptions~\cite{almeida2026jev}.
Recent studies explore candidate probability readout, decision-distribution distillation, and batched decisions to leverage the decision-making capabilities of pretrained models~\cite{li2026llmasjev,geuter2026innerjev,duan2026generaldecision,yu2026visualjev}.
Such models have also been applied to GUI automation and navigation, where agents select or verify candidate actions or waypoints, or represent predefined mid-level actions as single tokens~\cite{zhang2026jevmobile,sheng2026navjev,liu2026memvln,wang2026vernav}. However, applying decision models to VLN requires more than single-step option selection: agents need to predict consecutive actions while maintaining short-horizon temporal dependencies.
Existing robotic policies explore autoregressive action tokens and compressed action representations~\cite{kim2025openvla,pertsch2025fast}, as well as action chunking and parallel action prediction~\cite{zhao2023act,kim2025openvlaoft}.
Diffusion and flow models further generate continuous action chunks, but their original implementations require iterative sampling~\cite{chi2023diffusionpolicy,black2025pi0}.
In contrast, we predict multiple primitive-action distributions in parallel from a shared vision-language state representation, using temporally causal interactions among prediction features to capture short-horizon dependencies.
\section{Our Approach} \label{sec:method}
% Body fragment: the parent document provides the top-level method heading.
% Method overview figure (single column).
% Place this file and fig2(1).pdf in the project's fig/ directory.
% Insert with \input{fig/fig2}; cite with \figref{method-overview}.
\begin{figure}[t]
    \centering
    \includegraphics[width=\columnwidth]{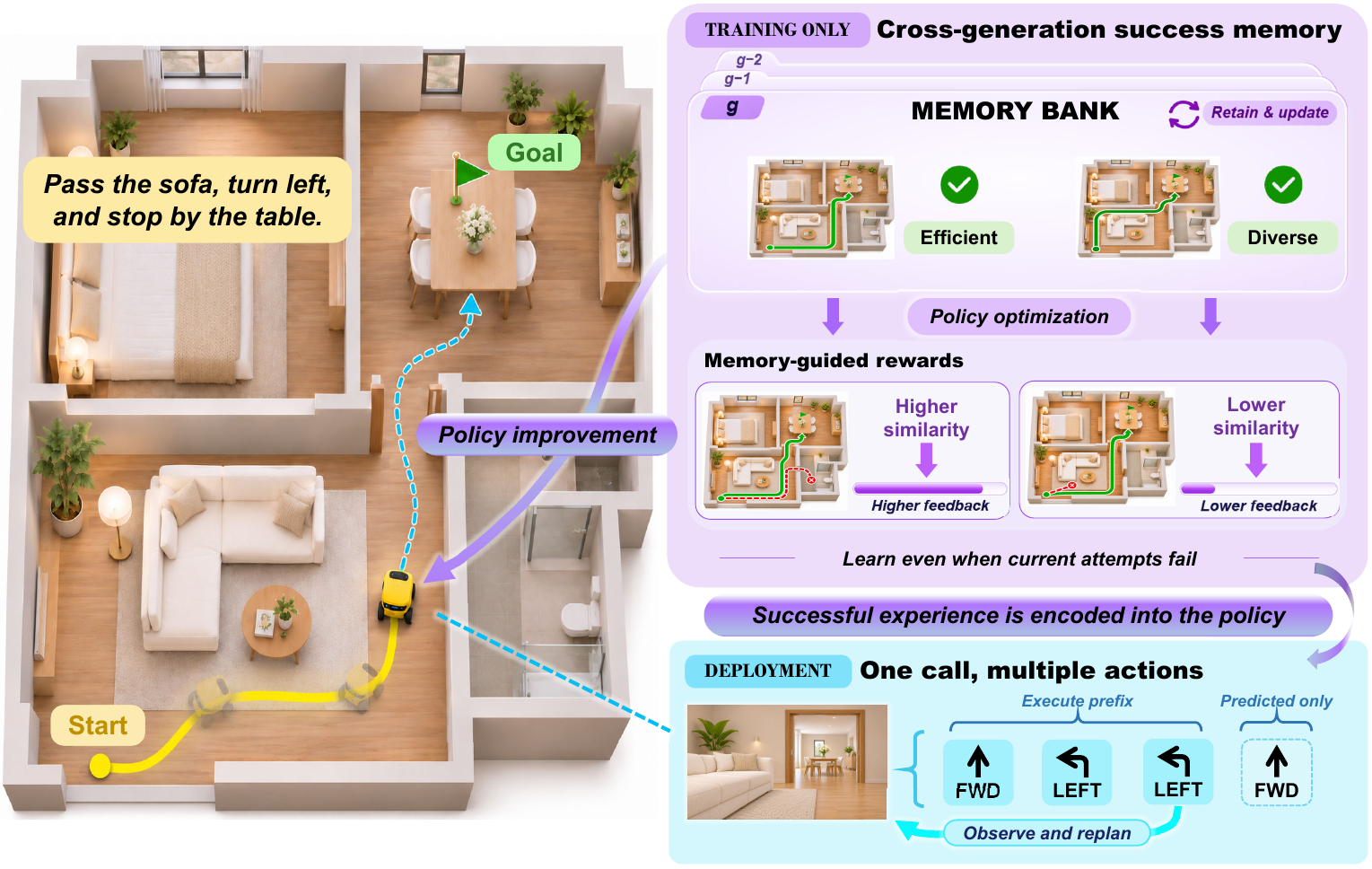}
    \caption{\textbf{Overview of NavDecision.}
During training, M-SRPO constructs trajectory-level rewards using current and historical successful trajectories encoded by a frozen world model. At deployment, PMAP predicts multiple action distributions in parallel, enabling multi-step execution with periodic replanning.}
    \label{fig:method-overview}
    \vspace{-3mm}
\end{figure}

% % Method overview figure (double-column).
% % Place this file and fig2.pdf in the project's fig/ directory.
% % Insert with \input{fig/fig2}; cite with \figref{method-overview}.
% \begin{figure*}[t]
%     \centering
%     \includegraphics[width=\textwidth]{fig/pdf/fig2.pdf}
%     \caption{\textbf{Overview of the proposed framework.} During training, M-SRPO constructs trajectory-level rewards from current and memorized successes in the representation space of a frozen world-model encoder. Memory retains path-efficient, representationally diverse trajectories. At deployment, PMAP predicts multiple action distributions in one pass; the agent executes a prefix and replans from a new observation.}
%     \label{fig:method-overview}
% \end{figure*}

As shown in \figref{method-overview}, we propose a multi-step decision model for vision-and-language navigation and a cross-iteration self-referential policy optimization framework. From a shared vision-language state representation, the model predicts action distributions in parallel through temporally causal interactions among prediction features. During reinforcement learning, current and historical successful trajectories jointly define a trajectory-level reward for training the model, allowing long-horizon navigation experience to improve the local action policy. The two components support consecutive action execution over short horizons and alignment with long-horizon goals, respectively. \secref{task-formulation} defines the navigation task and the decision model. \secref{parallel-multi-step-action-prediction} presents parallel multi-step action prediction and its supervised training, and \secref{cross-iteration-self-referential-policy-optimization} introduces cross-iteration self-referential policy optimization.

\subsection{Task Formulation}
\label{sec:task-formulation}

\Paragraph{Vision-and-language navigation.} Given a natural-language instruction $I$, an agent repeatedly selects actions from visual observations to reach the specified goal and stop~\cite{krantz2020vlnce}. At decision step $t$, let $\mathcal O_t$ denote the available current and historical visual observations, and let $\mathcal H_t$ denote the history of executed primitive actions. The decision context is $x_t=(I,\mathcal O_t,\mathcal H_t)$, and the primitive action space is $\mathcal A=\{\mathrm{STOP},\mathrm{FORWARD},\mathrm{LEFT},\mathrm{RIGHT}\}$. Here, $t$ indexes model decisions, each of which may execute multiple primitive actions. A VLM with an autoregressive action interface formulates navigation action output as conditional text generation: given $x_t$, it generates an action response token by token and parses the response into one or more executable actions~\cite{cheng2025navila,wei2025streamvln}. Let $\mathbf y_t=(y_{t,1},\ldots,y_{t,N_t})$ denote a response with $N_t$ tokens. Its probability factorizes as
\begin{equation}
p_{\mathrm{AR}}(\mathbf y_t\mid x_t)
=\prod_{n=1}^{N_t}
p_{\mathrm{AR}}(y_{t,n}\mid x_t,\mathbf y_{t,<n}),
\label{eq:ar-token-factorization}
\end{equation}
Here, $n$ indexes tokens within the response, and $\mathbf y_{t,<n}$ is the generated token prefix. An action description may contain multiple tokens, each conditioned on the preceding prefix. Producing a parsable action therefore requires sequential token-by-token decoding.

\Paragraph{Single-step decision model.} A decision interface predicts probabilities over candidate options from an input state, a decision question, and the options~\cite{almeida2026jev,li2026llmasjev}. For VLN, the input is $x_t$, the question is ``Which action should be executed now to follow the navigation instruction?'', and the candidate options are the primitive actions in $\mathcal A$. The model outputs their conditional probabilities. We instantiate this interface as an action classification model built on a VLM: the VLM encodes $x_t$, and an action head produces a score vector $\mathbf z_\theta(x_t)\in\mathbb R^{|\mathcal A|}$, yielding the navigation policy
\begin{equation}
\pi_\theta(\cdot\mid x_t)
=\operatorname{softmax}\!\left(\mathbf z_\theta(x_t)\right),
\label{eq:single-step-action-policy}
\end{equation}
Here, $\theta$ denotes the trainable policy parameters, and softmax normalizes over the candidate action set. Under the same navigation context and action space, this interface directly predicts probabilities over executable actions.

\subsection{Parallel Multi-Step Action Prediction}
\label{sec:parallel-multi-step-action-prediction}

Although single-step decision models remove token-by-token decoding, each model call outputs only the current primitive-action distribution. They do not explicitly model short-horizon action sequences, and consecutive action execution requires repeated model calls. We extend the model to parallel multi-step action prediction with a prediction horizon of $K$. Each position predicts over $\mathcal A$, and the action combination is generated from the current context. Multiple actions share a single backbone computation, while temporally causal interactions among prediction features capture short-horizon temporal dependencies.

\begin{figure}[t]
    \centering
    \includegraphics[width=0.95 \columnwidth]{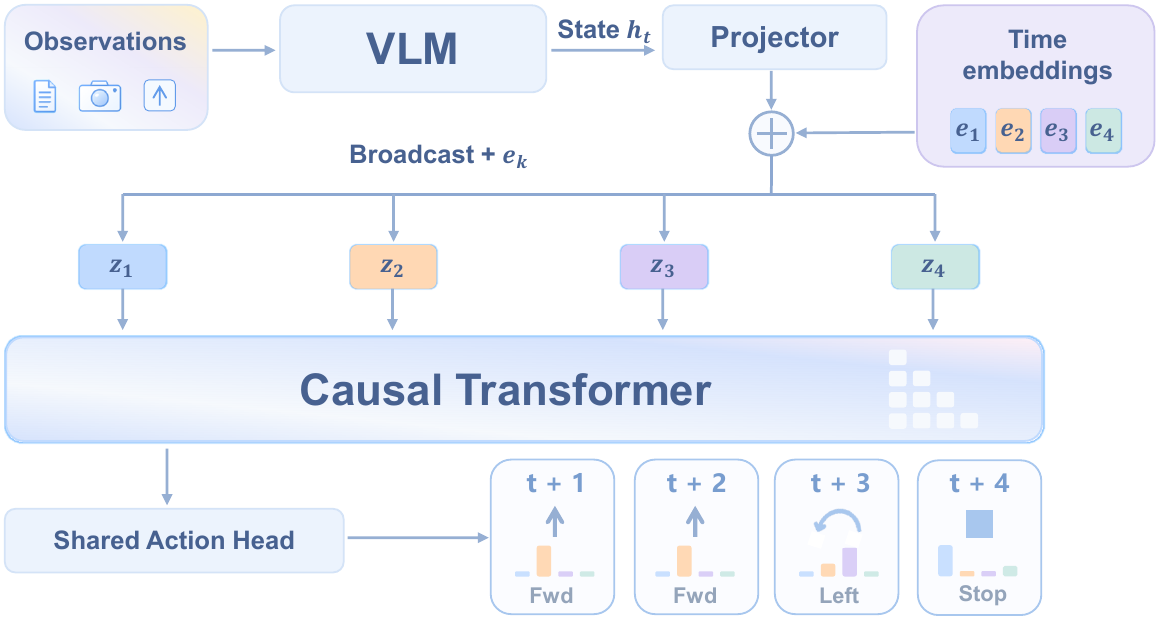}
    \caption{\textbf{Temporally causal action head for PMAP.}
The shared vision-language state is combined with learnable temporal embeddings. A causal Transformer and shared action head predict all action distributions in parallel.}
    \label{fig:pmap-head}
    \vspace{-3mm}
\end{figure}

\Paragraph{Temporally causal action head.} As shown in \figref{pmap-head}, at decision step $t$, we use the hidden representation of the last prompt token from the VLM backbone as the shared vision-language state representation $\mathbf h_t$. For prediction position $k\in\{0,\ldots,K-1\}$, we project this representation into the action feature space and add a learnable temporal embedding $\mathbf e_k$. A causal Transformer allows the features to interact across positions, and a shared classification layer produces the action distributions:
\begin{equation}
\begin{aligned}
\mathbf u_{t,k}^{(0)}
&=\operatorname{Proj}(\mathbf h_t)+\mathbf e_k,\\
\mathbf U_t
&=\operatorname{CausalTr}(\mathbf U_t^{(0)}),\\
\pi_\theta(\cdot\mid x_t,k)
&=\operatorname{softmax}\!\left(
\operatorname{Head}(\mathbf u_{t,k})\right),
\end{aligned}
\label{eq:temporally-causal-action-head}
\end{equation}
Here, $\operatorname{Proj}$ is a shared affine projection, $\mathbf U_t^{(0)}$ concatenates the input features along the temporal dimension, and $\mathbf u_{t,k}$ is the feature at position $k$ in the output matrix $\mathbf U_t$. $\operatorname{Head}$ applies layer normalization followed by an affine classification layer shared across positions to obtain action scores. Softmax normalizes along the action-class dimension. Causal self-attention allows position $k$ to read only prediction features at positions $j\le k$, modeling temporal dependencies among prediction positions. All action distributions are obtained in a single forward pass, and actions are sampled independently across positions conditioned on $x_t$. Temporal interactions occur among prediction features; sampled actions are not inputs to subsequent positions within the window.

\Paragraph{Multi-step supervised training.}
For supervised fine-tuning (SFT), we construct action
supervision windows from expert trajectories. Let $x_b$
denote the context of sample $b$, and let $a^\star_{b,k}$
denote its expert action at position $k$. We train the
action head by minimizing the mean cross-entropy over all
valid action positions in the constructed supervision
windows, denoted by $\Omega$:
\begin{equation}
\mathcal L_{\mathrm{SFT}}(\theta)
=-\frac{1}{|\Omega|}
\sum_{(b,k)\in\Omega}
\log\pi_\theta(a^\star_{b,k}\mid x_b,k),
\label{eq:multi-step-action-supervision}
\end{equation}
The set $\Omega$ includes valid actions up to and including the first STOP, excluding subsequent positions and those beyond the trajectory boundary. Multi-step prediction primarily models short-horizon temporal dependencies. To further align local decisions with long-horizon navigation goals, we introduce cross-iteration self-referential policy optimization. It improves the decision policy using trajectory-level rewards derived from historical successful trajectories, while retaining parallel multi-step action prediction at inference.

% PMAP action-head figure, for the second method subsection (single column).
% Place this file and fig3.pdf in the project's fig/ directory.
% Insert with \input{fig/fig3}; cite with \figref{pmap-head}.

\subsection{Memory-Augmented SRPO}
%  self-referential policy optimization 
\label{sec:cross-iteration-self-referential-policy-optimization}

With only terminal binary rewards, unsuccessful trajectories with different levels of progress receive identical feedback. Following the self-referential reward design of SRPO~\cite{fei2026srpo}, we evaluate unsuccessful trajectories against successful trajectories in the world-model representation space. We extend this design with cross-iteration success memory to continually reuse available references and provide trajectory-level training signals for the multi-step decision model.

\Paragraph{Cross-iteration success memory.} Success references drawn only from the current iteration are unavailable when that iteration contains no successful trajectories. This can occur in long-horizon navigation even when earlier exploration has already produced successful trajectories. We therefore maintain a cross-iteration success memory for each navigation task instance, using both current and historical successful trajectories as references. At training iteration $g$, the rollout policy generates $G$ trajectories for the same navigation task instance $\xi$. Let $\tau_i$, $s_i\in\{0,1\}$, and $L_i$ denote trajectory $i$, its environment-provided success indicator, and its total number of executed primitive actions, respectively. A frozen world-model encoder provides an L2-normalized trajectory embedding $\mathbf v_i$ for reward computation during training. We cluster valid embeddings of successful trajectories from the current iteration and denote the normalized current-iteration success centroids by $\mathcal C_g^{\mathrm{cur}}$. Let $\mathcal M_{g-1}$ be the success memory from the preceding iteration, initialized as $\mathcal M_0=\varnothing$. Each record $m$ stores $(\mathbf v_m,L_m,\mathrm{SPL}_m)$, where $\mathrm{SPL}_m$ measures the path efficiency of the successful trajectory~\cite{anderson2018evaluation}. The combined success reference set for the current iteration is
\begin{equation}
\mathcal C_g
=\mathcal C_g^{\mathrm{cur}}
\cup\{\mathbf v_m:m\in\mathcal M_{g-1}\},
\label{eq:combined-success-reference-set}
\end{equation}
Current successful embeddings form centroids through clustering. Historical records contribute embeddings and are not reclustered. To reuse experience from successful trajectories while controlling memory size, let $\mathcal S_g^+$ denote the valid successful records from the current iteration. We combine them with the historical memory into the candidate pool $\mathcal P_g=\mathcal M_{g-1}\cup\mathcal S_g^+$ and retain at most two records according to path efficiency and representational diversity:
\begin{equation}
\begin{aligned}
m^\star
&=\operatorname*{arg\,max}_{m\in\mathcal P_g}\mathrm{SPL}_m,\\
m^\dagger
&=\operatorname*{arg\,max}_{\substack{m\in\mathcal P_g\setminus\{m^\star\}\\L_m\le\gamma L_{m^\star}}}
\|\mathbf v_m-\mathbf v_{m^\star}\|_2,
\end{aligned}
\label{eq:cross-iteration-success-memory-selection}
\end{equation}
Here, $\gamma\ge1$ is the trajectory-length tolerance factor shared by memory selection and reward computation. The second record is retained only if a candidate exists and its embedding distance from the first record is at least $\delta_m>0$. Ties for $m^\star$ are resolved in favor of fewer primitive actions. Ties for $m^\dagger$ are resolved first by higher SPL and then by fewer primitive actions. The updated memory $\mathcal M_g$ is used from the next iteration onward and remains unchanged if no new valid successful records are obtained.

\Paragraph{Rewards from combined success references.} We combine trajectory length and the distance distribution to compute bounded similarity-based rewards. When $\mathcal C_g\ne\varnothing$, we use $L_{\mathrm{med}}^+$, the median trajectory length in primitive actions across current-iteration and historical successful trajectories, as the length reference. Let $\mathcal F$ contain the indices $i$ for which $s_i=0$, $\mathbf v_i$ is valid, and $L_i\le\gamma L_{\mathrm{med}}^+$, where $\gamma\ge1$ is the trajectory-length tolerance factor. For $i\in\mathcal F$, we compute the distance to the nearest success reference:
\begin{equation}
d_i
=\min_{\mathbf v\in\mathcal C_g}
\|\mathbf v_i-\mathbf v\|_2,
\label{eq:nearest-success-reference-distance}
\end{equation}
When $\mathcal F\ne\varnothing$, let $d_{\min}$ and $d_{\max}$ be the minimum and maximum distances within this set, and let $\Delta_d=d_{\max}-d_{\min}$. For $i\in\mathcal F$ with $\Delta_d\ge\delta_d>0$, define the normalized distance as $\widetilde d_i=(d_i-d_{\min})/\Delta_d$. The trajectory reward is
\begin{equation}
R_i=
\begin{cases}
1,&s_i=1,\\[2pt]
\alpha\operatorname{sigmoid}\!\left[\lambda\left(\tfrac12-\widetilde d_i\right)\right],
&i\in\mathcal F,\ \Delta_d\ge\delta_d,\\[3pt]
\alpha/2,&i\in\mathcal F,\ \Delta_d<\delta_d,\\[2pt]
0,&\text{otherwise},
\end{cases}
\label{eq:combined-reference-trajectory-reward}
\end{equation}
Here, $0<\alpha<1$ controls the reward scale for unsuccessful trajectories, $\lambda>0$ controls the steepness of the sigmoid, and $\delta_d$ is the distance-range threshold. This reward distinguishes eligible unsuccessful trajectories by their embedding distance to successful trajectory references and assigns every successful trajectory a higher reward. When the distance range is below $\delta_d$, assigning $\alpha/2$ avoids amplifying minor distance differences. When $\mathcal C_g=\varnothing$, we set $\mathcal F=\varnothing$. When $\mathcal F=\varnothing$, no unsuccessful trajectory is available for self-referential evaluation, and we use the terminal binary success reward $R_i=s_i$.

\Paragraph{Policy optimization.}
We optimize the multi-step decision policy using the clipped policy objective in SRPO~\cite{fei2026srpo}, following GRPO~\cite{shao2024deepseekmath}. At each decision step $t$ of trajectory $i$, the policy predicts $K$ actions in parallel and randomly samples an execution cap $E_{i,t}\in\{1,\ldots,K\}$. Actions are executed sequentially until the cap is reached, STOP is executed, or the environment terminates. Let $\ell_{i,t}$ denote the number of executed actions, forming the valid action prefix. The same execution-cap sampling mechanism is used during training and rollout.

We normalize trajectory-level rewards within each task group to obtain a shared advantage $\widehat A_i$ for all decisions in trajectory $i$. Policy updates are performed over the executed action prefixes using a joint importance ratio:
\begin{equation}
\rho_{i,t}
=
\prod_{k=0}^{\ell_{i,t}-1}
\frac{
\pi_\theta(a_{i,t,k}\mid x_{i,t},k)
}{
\pi_{\mathrm{old}}(a_{i,t,k}\mid x_{i,t},k)
}.
\label{eq:executed-prefix-importance-ratio}
\end{equation}

We apply PPO-style clipping~\cite{schulman2017ppo} and KL regularization against the initial multi-step SFT policy, following the optimization objective of SRPO. The objective averages valid decision positions within each trajectory before averaging across trajectories, assigning equal weight to each trajectory. The RL objective is
\begin{equation}
\begin{aligned}
\mathcal{L}_{\mathrm{RL}}(\theta)
=&\frac{1}{|\mathcal{B}|}
\sum_{\tau_i\in\mathcal{B}}
\frac{1}{|\mathcal{J}_i|}
\sum_{t\in\mathcal{J}_i}
\Big[
-\min\big\{
\rho_{i,t}\widehat A_i,\,
\rho_{i,t}^{\mathrm{clip}}\widehat A_i
\big\}\\
&\qquad\qquad
+\beta D_{i,t}
\Big],
\end{aligned}
\label{eq:rl-objective}
\end{equation}
% Trajectories with negligible group reward variance receive zero advantages. Newly obtained successful trajectories are added to the cross-iteration success memory for subsequent training iterations. Detailed derivations of the objective and the final RL loss are provided in the supplementary material.
Here, $\mathcal{B}$ is the trajectory minibatch, and
$\mathcal{J}_i$ denotes the valid decision positions in
trajectory $i$. $\rho_{i,t}^{\mathrm{clip}}$ is the clipped
joint importance ratio, and $D_{i,t}$ denotes the KL
divergence summed over executed action positions, with
$\beta$ controlling its regularization strength.
\section{Experiments} \label{sec:exp}
% English experiment body fragment, including all five main tables.
% The parent document supplies \section{Experiments}.
% Add \input{experiment_preamble} in the parent preamble if the packages/macros
% used here are not already available. Keep figures/ at the project root.
% Supplementary cross-references resolve when appendix.tex (or the inline
% appendix_experiments.tex) from the accompanying project is included.
\label{sec:experiments}

\begin{table*}[t]
\centering
% BEGIN tables/main_results_content
\caption{\textbf{Navigation results on R2R-CE and RxR-CE \texttt{val\_unseen}.}
Bold indicates the best result for each metric within each method group.}
\label{tab:main}
{\footnotesize
\setlength{\tabcolsep}{2.7pt}
\renewcommand{\arraystretch}{1.08}
\resizebox{\textwidth}{!}{%
\begin{threeparttable}
\begin{tabular}{@{}lcc cccc rrrr rrrr@{}}
\toprule
\multirow{2}{*}{\textbf{Method}}
& \multirow{2}{*}{\shortstack{\textbf{Backbone}\\\textbf{size}}}
& \multirow{2}{*}{\shortstack{\textbf{Direct}\\\textbf{action}}}
& \multicolumn{4}{c}{\textbf{Observation}}
& \multicolumn{4}{c}{\textbf{R2R-CE}}
& \multicolumn{4}{c}{\textbf{RxR-CE}} \\
\cmidrule(lr){4-7}\cmidrule(lr){8-11}\cmidrule(l){12-15}
& &
& Pano. & Odo. & Depth & S.RGB
& \cellcolor{red!12}NE$\downarrow$
& \cellcolor{gray!15}OSR$\uparrow$
& \cellcolor{gray!15}SR$\uparrow$
& \cellcolor{gray!15}SPL$\uparrow$
& \cellcolor{red!12}NE$\downarrow$
& \cellcolor{gray!15}SR$\uparrow$
& \cellcolor{gray!15}SPL$\uparrow$
& \cellcolor{gray!15}nDTW$\uparrow$ \\
\midrule

\multicolumn{15}{@{}l}{\textit{Conventional navigation methods}} \\

HPN+DN~\cite{krantz2021waypoint}
& \na & \na
& $\checkmark$ & $\checkmark$ & $\checkmark$ &
& 6.31 & 40.00 & 36.00 & 34.00
& \na & \na & \na & \na \\

CMA~\cite{georgakis2022crossmodal}
& \na & \na
& $\checkmark$ & $\checkmark$ & $\checkmark$ &
& 6.20 & 52.00 & 41.00 & 36.00
& 8.76 & 26.60 & 22.20 & 47.10 \\

ETPNav~\cite{an2025etpnav}
& \na & \na
& $\checkmark$ & $\checkmark$ & $\checkmark$ &
& 4.71 & 65.00 & 57.00 & 49.00
& \textbf{5.64} & 54.80 & 44.90 & 61.90 \\

NavMorph~\cite{yao2025navmorph}
& \na & \na
& & & $\checkmark$ & $\checkmark$
& 5.75 & 56.90 & 47.90 & 33.20
& 8.85 & 30.80 & 22.80 & 44.20 \\

SC$^2$-WM~\cite{yao2026sc2wm}
& \na & \na
& $\checkmark$ & $\checkmark$ & $\checkmark$ &
& \textbf{4.25} & \textbf{70.00} & \textbf{66.00} & \textbf{54.00}
& 5.72 & \textbf{60.00} & \textbf{50.00} & \textbf{67.00} \\

\midrule
\multicolumn{15}{@{}l}{\textit{VLM-based navigation methods}} \\

NaVid~\cite{zhang2024navid}
& 7B & $\times$
& & & & $\checkmark$
& 5.47 & 49.10 & 37.40 & 35.90
& \na & \na & \na & \na \\

Uni-NaVid~\cite{zhang2025uninavid}
& 7B & $\times$
& & & & $\checkmark$
& 5.58 & 53.30 & 47.00 & 42.70
& 6.24 & 48.70 & 40.90 & \na \\

NaVILA~\cite{cheng2025navila}
& 8B & $\times$
& & & & $\checkmark$
& 5.22 & 62.50 & 54.00 & 49.00
& 6.77 & 49.30 & 44.00 & 58.80 \\

ActiveVLN~\cite{zhang2026activevln}
& 3B & $\times$
& & & & $\checkmark$
& 5.31 & 58.30 & 50.10 & 43.70
& 5.84 & 50.70 & 41.20 & 58.10 \\

StreamVLN~\cite{wei2025streamvln}
& 7B & $\times$
& & & & $\checkmark$
& 4.98 & 64.20 & 56.90 & 51.90
& 6.22 & 52.90 & 46.00 & 61.90 \\

NaVIDA~\cite{zhu2026navida}
& 3B & $\times$
& & & & $\checkmark$
& 4.32 & 69.50 & 61.40 & 54.70
& 5.23 & 57.40 & 49.60 & 67.00 \\

InternVLA-N1 (S2)~\cite{wei2026dualvln}
& 7B & $\times$
& & & & $\checkmark$
& 4.89 & 60.60 & 55.40 & 52.10
& 6.41 & 49.50 & 41.80 & 62.60 \\

InternVLA-N1 (S1+S2)~\cite{wei2026dualvln}
& 7B & $\times$
& & & $\checkmark$ & $\checkmark$
& 4.83 & 63.30 & 58.20 & 54.00
& 5.91 & 53.50 & 46.10 & 65.30 \\

NavFoM~\cite{zhang2026navfom}
& 7B & $\times$
& & & & $\checkmark$
& 5.01 & 64.90 & 56.20 & 51.20
& 5.51 & 57.40 & 49.40 & 60.20 \\

JanusVLN~\cite{zeng2026janusvln}
& 7B & $\times$
& & & & $\checkmark$
& 4.78 & 65.20 & 60.50 & 56.80
& 6.06 & 56.20 & 47.50 & 62.10 \\

DualVLN~\cite{wei2026dualvln}
& 7B & $\times$
& & & & $\checkmark$
& 4.05 & 70.70 & 64.30 & 58.50
& 4.58 & 61.40 & 51.80 & \textbf{70.00} \\

AwareVLN~\cite{guo2026awarevln}
& 8B & $\times$
& & & & $\checkmark$
& \textbf{4.02} & 73.50 & 65.40 & 55.10
& \textbf{3.95} & 67.60 & 56.10 & 65.70 \\

FutureNav-4B~\cite{zhang2026futurenav}
& 4B & $\times$
& & & & $\checkmark$
& 4.24 & 70.20 & 65.40 & \textbf{61.30}
& 4.60 & 60.90 & 52.80 & 68.00 \\

FutureNav-8B~\cite{zhang2026futurenav}
& 8B & $\times$
& & & & $\checkmark$
& 4.29 & 69.20 & 64.30 & 60.30
& 4.26 & 63.90 & 54.80 & 69.10 \\

Qwen-RobotNav-4B~\cite{qwenrobotnav2026}
& 4B & $\times$
& & & & $\checkmark$
& 4.22 & \textbf{73.60} & \textbf{66.90} & 60.50
& 4.15 & 71.30 & 61.50 & 68.60 \\

Qwen-RobotNav-8B~\cite{qwenrobotnav2026}
& 8B & $\times$
& & & & $\checkmark$
& 4.36 & 72.70 & 65.70 & 59.60
& 4.16 & \textbf{73.40} & \textbf{63.50} & 69.90 \\

\midrule
\multicolumn{15}{@{}l}{\textit{Decision-based navigation methods}} \\

OmniJev~\cite{almeida2026jev}
& 4B & $\checkmark$
& & & & $\checkmark$
& 9.47 & 3.75 & 1.47 & 1.30
& 11.95 & 8.07 & 6.97 & 26.59 \\

\rowcolor{gray!5}
NavDecision ($E=1$)
& 4B & $\checkmark$
& & & & $\checkmark$
& 4.12 & 76.89 & 64.33 & \textbf{55.31}
& \textbf{4.00} & \textbf{65.03} & \textbf{54.38} & \textbf{68.35} \\

\rowcolor{gray!15}
NavDecision ($E=2$)
& 4B & $\checkmark$
& & & & $\checkmark$
& \textbf{3.99} & \textbf{77.98} & \textbf{64.60} & 54.31
& 4.05 & 64.57 & 52.84 & 67.59 \\

\rowcolor{gray!5}
NavDecision ($E=4$)
& 4B & $\checkmark$
& & & & $\checkmark$
& 4.34 & 76.35 & 60.79 & 48.72
& 4.50 & 60.81 & 48.11 & 63.30 \\

\bottomrule
\end{tabular}

\begin{tablenotes}[flushleft]
\footnotesize
\item Pano.: panoramic RGB; Odo.: odometry; Depth: depth input;
S.RGB: single-view forward-facing RGB.
Observation inputs are marked with $\checkmark$ and left blank otherwise.
Direct action indicates discrete action prediction through an action head
($\checkmark$/$\times$).
\end{tablenotes}

\end{threeparttable}}}
% END tables/main_results_content
\end{table*}
% END tables/main_results

\subsection{Experimental Setup}
\Paragraph{Datasets and metrics.} We evaluate our method on the R2R-CE~\cite{krantz2020vlnce} and RxR-CE~\cite{ku2020rxr} val unseen splits. The policy is first pre-trained via SFT on R2R~\cite{anderson2018vln}, RxR~\cite{ku2020rxr}, ScaleVLN~\cite{wang2023scalevln} and EnvDrop~\cite{tan2019envdrop}, followed by online reinforcement learning on the R2R and RxR training tasks. We adopt SR and SPL as primary metrics, while also reporting NE, OSR, and nDTW~\cite{anderson2018evaluation}. Further details are provided in the supplementary material.

\Paragraph{Model and training.} Our policy, built on OmniJev with a Qwen3.5-4B backbone, takes instructions, current s.RGB images, and the 8 most recent observations and actions as input. SFT trains only the LoRA adapters and the action head. Online training samples 8 complete trajectories per group. A frozen V-JEPA 2~\cite{assran2025vjepa2} ViT-g encodes these trajectories, with at most two successful records retained per task. Both the trajectory encoder and success memory are utilized exclusively during the training phase.

\Paragraph{Execution settings.} The primitive actions are STOP, moving forward by 0.25 m, and turning left or right by $15^\circ$. Let $K$ be the number of actions predicted per model call and $E$ the execution cap before acquiring a new observation. Execution ends early at STOP or environment termination. We fix $K=4$, compare $E\in\{1,2,4\}$ in the main results, and use $E=2$ by default for ablations. Evaluation selects the highest-probability action at each prediction position.

\subsection{Main Results}
\label{sec:comparison}

\Paragraph{Navigation performance.}
Our direct multi-step policy achieves competitive navigation success
on both benchmarks (Table~\ref{tab:main}).
With the default execution cap $E=2$, our method obtains
64.60\% SR and 54.31\% SPL on R2R-CE, and
64.57\% SR and 52.84\% SPL on RxR-CE.
NavDecision achieves performance comparable to the strong autoregressive baseline FutureNav-4B~\cite{zhang2026futurenav}, with only a 0.80\% lower SR on R2R-CE and a 3.67\% higher SR on the more challenging RxR-CE benchmark. These results underscore that our parallel decision framework preserves robust navigation capabilities without the prohibitive computational overhead of token-by-token decoding.

\Paragraph{Inference efficiency.}
The model achieves these competitive success rates while significantly reducing inference latency (Table~\ref{tab:latency}).
With $E=2$, the amortized model inference time per executed
primitive action is 55.10~ms on R2R-CE and 62.17~ms on RxR-CE.
This yields approximately $3.3\times$ speedups even over
the fastest timed baseline on each benchmark.
Compared with FutureNav-4B~\cite{zhang2026futurenav}, the corresponding speedups are
$11.06\times$ and $9.50\times$, respectively, while retaining
the competitive success rates discussed above. 
These significant speedups are primarily attributed to PMAP, which eliminates token-by-token decoding latency and maximizes computational efficiency by sharing a single backbone pass across multiple actions. 

\Paragraph{Performance--efficiency trade-off.}
With the prediction horizon fixed at $K=4$, the execution cap
$E$ controls how frequently the policy acquires observations
and replans.
Moving from $E=1$ to $E=2$ approximately halves the amortized
inference time, while changing SR by less than 0.5 percentage
points on either benchmark.
Setting $E=4$ further reduces
the amortized inference time to approximately 30~ms per
primitive action, achieving around $10\times$ speedup
over the median inference time of the evaluated
autoregressive baselines. This comes at the cost of
approximately 3.8 percentage points in SR compared
with $E=2$ on both benchmarks, offering a substantially
faster execution mode with a moderate
performance trade-off.

% % BEGIN tables/overall
% \begin{table}[t]
% \centering
% \caption{\textbf{Overall ablation.} Baseline uses single-step SFT ($K=E=1$). PMAP uses multi-step SFT, followed by SRPO~\cite{fei2026srpo} or M-SRPO ($K=4,E=2$).}
% \label{tab:overall}
% {\small
% \setlength{\tabcolsep}{8pt}
% \renewcommand{\arraystretch}{1.1}
% \begin{threeparttable}
% \begin{tabularx}{\columnwidth}{@{}Xrrrr@{}}
% \toprule
% \multirow{2}{*}{\textbf{Configuration}} & \multicolumn{2}{c}{\textbf{R2R-CE}} & \multicolumn{2}{c}{\textbf{RxR-CE}} \\
% \cmidrule(lr){2-3}\cmidrule(l){4-5}
%  & SR$\uparrow$ & SPL$\uparrow$ & SR$\uparrow$ & SPL$\uparrow$ \\
% \midrule
% Baseline & 46.55 & 41.95 & 51.40 & 40.92 \\
% PMAP & 47.96 & 39.79 & 53.69 & 43.28 \\
% PMAP+SRPO & 62.26 & 52.70 & 61.92 & 50.82 \\
% \rowcolor{gray!15}
% PMAP+M-SRPO & 64.60 & 54.31 & 64.57 & 52.84 \\
% \bottomrule
% \end{tabularx}
% \begin{tablenotes}[flushleft]
% \footnotesize
% \item SFT uses the same four data sources. PMAP+M-SRPO corresponds to $E=2$ in Table~\ref{tab:main}.
% \end{tablenotes}
% \end{threeparttable}}
% \end{table}

% % END tables/overall

% BEGIN tables/overall

% END tables/overall

\subsection{Ablation Studies}
\label{sec:ablation}

We evaluate the contributions of parallel multi-step action
prediction (PMAP) and memory-augmented self-referential
policy optimization (M-SRPO).

% PMAP action-head ablation and efficiency figure (three panels, two columns).
% Place this file and structure_ablation_cvpr(1).pdf in fig/.
% Insert with \input{fig/fig4}; the existing \ref{fig:structure} remains valid.
% Replace the original structure figure block rather than adding a second copy.
\begin{figure*}[t]
    \centering
    \includegraphics[width=0.9\textwidth]{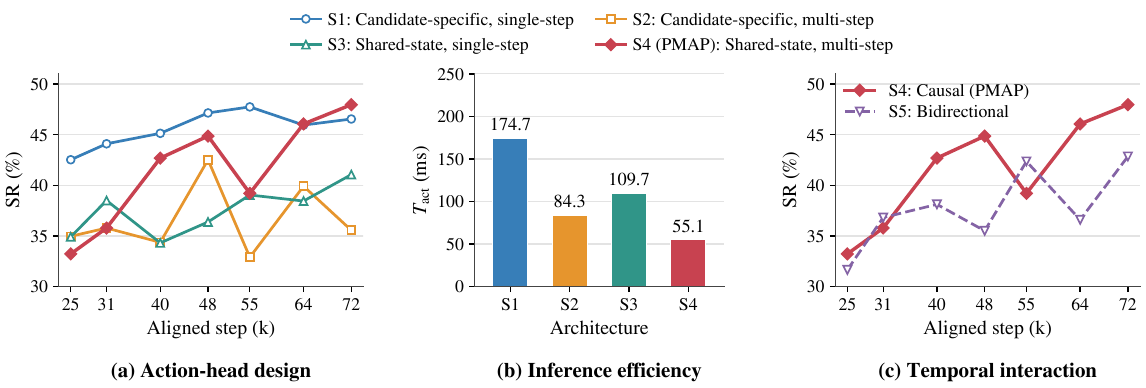}
    \caption{\textbf{PMAP ablation and inference efficiency.}
(a) SFT curves of S1--S4 on R2R-CE \texttt{val\_unseen};
(b) inference time per action;
(c) causal versus bidirectional temporal attention (S4 vs. S5).
S1--S4 compare candidate-specific/shared-state and single-/multi-step designs.}
    \label{fig:structure}
\end{figure*}

\Paragraph{Overall ablation.}
Table~\ref{tab:overall} compares the single-step SFT
Baseline, PMAP, PMAP+SRPO, and the full PMAP+M-SRPO,
isolating the effects of multi-step prediction and
trajectory-level policy optimization.

PMAP improves the success rate on both benchmarks,
with gains of 1.41 and 2.29 percentage points on R2R
and RxR, respectively. More importantly, adding
trajectory-level policy optimization substantially
boosts navigation performance. The full PMAP+M-SRPO
achieves 64.60\% SR and 54.31\% SPL on R2R, and
64.57\% SR and 52.84\% SPL on RxR. Compared with
PMAP, it improves SR by 16.64 points on R2R and
10.88 points on RxR. Overall, the full method
outperforms the SFT Baseline by 18.05 and 13.17
SR points on R2R and RxR, respectively, highlighting
the effectiveness of trajectory-level reinforcement
learning for long-horizon navigation.

\begin{table}[t]
\centering
\caption{\textbf{Navigation performance and inference efficiency.}
All methods with publicly available weights are timed on an H20
under the same settings.
$T_{\mathrm{act}}$ denotes amortized model inference time per
executed primitive action (ms). Bold indicates the lowest
$T_{\mathrm{act}}$ on each benchmark.}
\label{tab:latency}
{\footnotesize
\setlength{\tabcolsep}{3pt}
\renewcommand{\arraystretch}{1.08}
\begin{threeparttable}
\begin{tabularx}{\columnwidth}{>{\raggedright\arraybackslash}Xcrrrr}
\toprule
\multirow{2}{*}{\textbf{Method}}
& \multirow{2}{*}{\shortstack{\textbf{Backbone}\\\textbf{size}}}
& \multicolumn{2}{c}{\textbf{R2R-CE}}
& \multicolumn{2}{c}{\textbf{RxR-CE}} \\
\cmidrule(lr){3-4}\cmidrule(lr){5-6}
& &
\cellcolor{gray!15}SR$\uparrow$
& \cellcolor{red!12}$T_{\mathrm{act}}\downarrow$
& \cellcolor{gray!15}SR$\uparrow$
& \cellcolor{red!12}$T_{\mathrm{act}}\downarrow$ \\
\midrule

NaVid~\cite{zhang2024navid}
& 7B & 37.40 & 343.47 & \na & 407.49 \\

Uni-NaVid~\cite{zhang2025uninavid}
& 7B & 47.00 & 229.07 & 48.70 & 234.93 \\

NaVILA~\cite{cheng2025navila}
& 8B & 54.00 & 858.18 & 49.30 & 838.81 \\

ActiveVLN~\cite{zhang2026activevln}
& 3B & 50.10 & 181.31 & 50.70 & 221.14 \\

StreamVLN~\cite{wei2025streamvln}
& 7B & 56.90 & 208.30 & 52.90 & 206.25 \\

NaVIDA~\cite{zhu2026navida}
& 3B & 61.40 & 277.60 & 57.40 & 290.91 \\

\mbox{InternVLA-N1 (S2)}~\cite{wei2026dualvln}
& 7B & 55.40 & 299.29 & 49.50 & 298.54 \\

\mbox{InternVLA-N1 (S1+S2)}~\cite{wei2026dualvln}
& 7B & 58.20 & 306.07 & 53.50 & 306.08 \\

JanusVLN~\cite{zeng2026janusvln}
& 7B & 60.50 & 2445.85 & 56.20 & 2455.48 \\

DualVLN~\cite{wei2026dualvln}
& 7B & 64.30 & 302.09 & 61.40 & 301.18 \\

AwareVLN~\cite{guo2026awarevln}
& 8B & 65.40 & 833.40 & 67.60 & 926.96 \\

FutureNav-4B~\cite{zhang2026futurenav}
& 4B & 65.40 & 609.49 & 60.90 & 590.92 \\

\midrule
\rowcolor{gray!5}
NavDecision ($E=1$)
& 4B & 64.33 & 109.49 & 65.03 & 124.96 \\

\rowcolor{gray!15}
NavDecision ($E=2$)
& 4B & 64.60 & 55.10 & 64.57 & 62.17 \\

\rowcolor{gray!5}
NavDecision ($E=4$)
& 4B & 60.79 & \textbf{27.37} & 60.81 & \textbf{31.17} \\

\bottomrule
\end{tabularx}
\end{threeparttable}
}
\end{table}

\subsubsection{PMAP}
\label{sec:pmap-ablation}

\Paragraph{Shared-state representation.}
OmniJev encodes each action candidate separately by appending its text to
a shared prefix containing observations, instructions, and action
history. Although the branches share backbone parameters and
prefix computation, their candidate representations remain
independent. We instead adopt a shared-state representation,
using the hidden state of the last prefix token to score all
primitive actions through a unified action head.

\Paragraph{Causal multi-step prediction.}
We introduce learnable temporal embeddings and a causal
Transformer to model dependencies among prediction features.
As shown in Figure~\ref{fig:structure}(a,c) and
Table~\ref{tab:structure}, PMAP (S4) achieves the highest
SFT success rate of 47.96\% on R2R-CE.
Compared with shared-state single-step prediction (S3),
it improves SR and SPL by 5.11 and 3.69 percentage points,
respectively. It also outperforms candidate-specific
multi-step prediction (S2) by 5.44 points in SR and
5.98 points in SPL. Replacing causal with bidirectional
interaction (S5) reduces SR and SPL by 5.11 and
4.75 points, respectively, confirming the importance
of causal temporal modeling.

\Paragraph{Inference efficiency.}
Figure~\ref{fig:structure}(b) shows that shared-state
prediction reduces model inference time per action from
174.71 to 109.72~ms, a 37.20\% reduction.
With $E=2$, PMAP further reduces amortized inference
time to 55.10~ms per executed primitive action,
compared with 84.27~ms for candidate-specific
multi-step prediction (S2). Overall, PMAP achieves
a $3.17\times$ speedup over the single-step
candidate-specific baseline (S1).

\begin{table}[t]
\centering
\caption{\textbf{Overall ablation.}
Baseline uses single-step SFT ($K=E=1$).
PMAP uses multi-step SFT, followed by SRPO~\cite{fei2026srpo}
or M-SRPO ($K=4,E=2$).}
\label{tab:overall}
{\small
\setlength{\tabcolsep}{8pt}
\renewcommand{\arraystretch}{1.1}
\begin{threeparttable}
\begin{tabularx}{\columnwidth}{Xrrrr}
\toprule
\multirow{2}{*}{\textbf{Configuration}}
& \multicolumn{2}{c}{\textbf{R2R-CE}}
& \multicolumn{2}{c}{\textbf{RxR-CE}} \\
\cmidrule(lr){2-3}\cmidrule(lr){4-5}
& SR$\uparrow$
& SPL$\uparrow$
& SR$\uparrow$
& SPL$\uparrow$ \\
\midrule
Baseline
& 46.55 & 41.95 & 51.40 & 40.92 \\
PMAP
& 47.96 & 39.79 & 53.69 & 43.28 \\
PMAP+SRPO
& 62.26 & 52.70 & 61.92 & 50.82 \\
\rowcolor{gray!15}
PMAP+M-SRPO
& 64.60 & 54.31 & 64.57 & 52.84 \\
\bottomrule
\end{tabularx}

\begin{tablenotes}[flushleft]
\footnotesize
\item SFT uses the same four data sources.
PMAP+M-SRPO corresponds to $E=2$ in Table~\ref{tab:main}.
\end{tablenotes}

\end{threeparttable}
}
\end{table}

% BEGIN tables/structure
\begin{table}[t]
\centering
\caption{\textbf{Ablation of action-head design on R2R-CE.}
All models are evaluated after SFT, without SRPO.}
\label{tab:structure}
{\footnotesize
\setlength{\tabcolsep}{4.8pt}
\renewcommand{\arraystretch}{1.12}
\begin{threeparttable}
\begin{tabularx}{\columnwidth}{lXcclrr}
\toprule
\multirow{2}{*}{\textbf{ID}}
& \multirow{2}{*}{\textbf{Representation}}
& \multirow{2}{*}{$K$}
& \multirow{2}{*}{$E$}
& \multirow{2}{*}{\shortstack{\textbf{Temporal}\\\textbf{interaction}}}
& \multicolumn{2}{c}{\textbf{R2R-CE}} \\
\cmidrule(lr){6-7}
& & & &
& SR$\uparrow$
& SPL$\uparrow$ \\
\midrule
S1 & Candidate-specific & 1 & 1 & \na
& 46.55 & 41.95 \\
S2 & Candidate-specific & 4 & 2 & Causal
& 42.52 & 33.81 \\
\midrule
S3 & Shared-state & 1 & 1 & \na
& 42.85 & 36.10 \\
\rowcolor{gray!15}
S4 & Shared-state & 4 & 2 & Causal
& 47.96 & 39.79 \\
S5 & Shared-state & 4 & 2 & Bidirectional
& 42.85 & 35.04 \\
\bottomrule
\end{tabularx}

\begin{tablenotes}[flushleft]
\footnotesize
\item Candidate-specific models retain candidate action branches;
shared-state models use a common context representation.
Single-step models have no temporal interaction across positions
(\na). S4 is PMAP.
\vspace{-2mm}
\end{tablenotes}

\end{threeparttable}
}
\end{table}
% END tables/structure

% % BEGIN tables/memory
% \begin{table}[t]
% \centering
% \caption{\textbf{M-SRPO ablation and GRPO comparison on R2R-CE.} Full metrics and checkpoint details are in Appendix.}
% \label{tab:memory}
% {\small
% \setlength{\tabcolsep}{3pt}
% \renewcommand{\arraystretch}{1.1}
% \begin{threeparttable}
% \begin{tabularx}{\columnwidth}{@{}Xlrrr@{}}
% \toprule
% \textbf{Policy} & \textbf{Optimization} & NE$\downarrow$ & SR$\uparrow$ & SPL$\uparrow$ \\
% \midrule
% Single-step & SRPO~\cite{fei2026srpo} & 4.62 & 58.95 & 51.64 \\
% Single-step & M-SRPO & 4.31 & 61.45 & 54.25 \\
% \midrule
% PMAP & GRPO~\cite{shao2024deepseekmath} & 4.86 & 56.23 & 47.69 \\
% PMAP & SRPO~\cite{fei2026srpo} & 4.29 & 62.26 & 52.70 \\
% \rowcolor{gray!15}
% PMAP & M-SRPO & 3.99 & 64.60 & 54.31 \\
% \bottomrule
% \end{tabularx}
% \begin{tablenotes}[flushleft]
% \footnotesize
% \item Single-step uses S1; PMAP uses S4. PMAP+M-SRPO is the full method.
% \end{tablenotes}
% \end{threeparttable}}
% \end{table}

% % END tables/memory

% BEGIN tables/memory
\begin{table}[t]
\centering
\caption{\textbf{M-SRPO and GRPO Ablation on R2R-CE.}}
\label{tab:memory}
{\small
\setlength{\tabcolsep}{8pt}
\renewcommand{\arraystretch}{1.1}
\begin{threeparttable}

\begin{tabularx}{\columnwidth}{Xlrrr}
\toprule
\multirow{2}{*}{\textbf{Policy}}
& \multirow{2}{*}{\textbf{Optimization}}
& \multicolumn{3}{c}{\textbf{R2R-CE}} \\
\cmidrule(lr){3-5}
&
& \cellcolor{red!12}NE$\downarrow$
& \cellcolor{gray!15}SR$\uparrow$
& \cellcolor{gray!15}SPL$\uparrow$ \\
\midrule

Single-step & SRPO~\cite{fei2026srpo}
& 4.62 & 58.95 & 51.64 \\

Single-step & M-SRPO
& 4.31 & 61.45 & 54.25 \\

\midrule

PMAP & GRPO~\cite{shao2024deepseekmath}
& 4.86 & 56.23 & 47.69 \\

PMAP & SRPO~\cite{fei2026srpo}
& 4.29 & 62.26 & 52.70 \\

\rowcolor{gray!15}
PMAP & M-SRPO
& 3.99 & 64.60 & 54.31 \\

\bottomrule
\end{tabularx}

\begin{tablenotes}[flushleft]
\footnotesize
\item Single-step uses S1; PMAP uses S4.
PMAP+M-SRPO is the full method.
\vspace{-2mm}
\end{tablenotes}

\end{threeparttable}
}
\end{table}
% END tables/memory

\subsubsection{M-SRPO}
\label{sec:msrpo-ablation}

M-SRPO extends SRPO with cross-iteration success memory,
allowing historical successful trajectories to provide
reward references across training iterations.
Table~\ref{tab:memory} evaluates its effectiveness with
single-step and multi-step policies, alongside a GRPO~\cite{shao2024deepseekmath} baseline.

Adding cross-iteration memory to SRPO improves the
single-step policy's SR by 2.50 percentage points and
SPL by 2.61 points. For PMAP, the corresponding gains
are 2.34 and 1.61 points, respectively.
Compared with GRPO~\cite{shao2024deepseekmath}, multi-step SRPO improves SR by
6.03 points and SPL by 5.01 points, while the full
PMAP+M-SRPO further increases these gains to 8.37
and 6.62 points.
Finally, combining M-SRPO with multi-step prediction
yields a 3.15-point SR improvement over its single-step
counterpart, with nearly identical SPL (54.31\% versus
54.25\%). These results demonstrate the complementary
benefits of cross-iteration success memory and multi-step
action prediction.

% Real-world navigation figure (single column).
% Copy realworld_navigation.png from the supplied project's figures/ to fig/.
% Insert with \input{fig/fig5}; the existing \ref{fig:realworld} remains valid.
% Replace the original real-world figure block rather than adding a second copy.
\begin{figure}[t]
    \centering
    \includegraphics[width=\columnwidth]{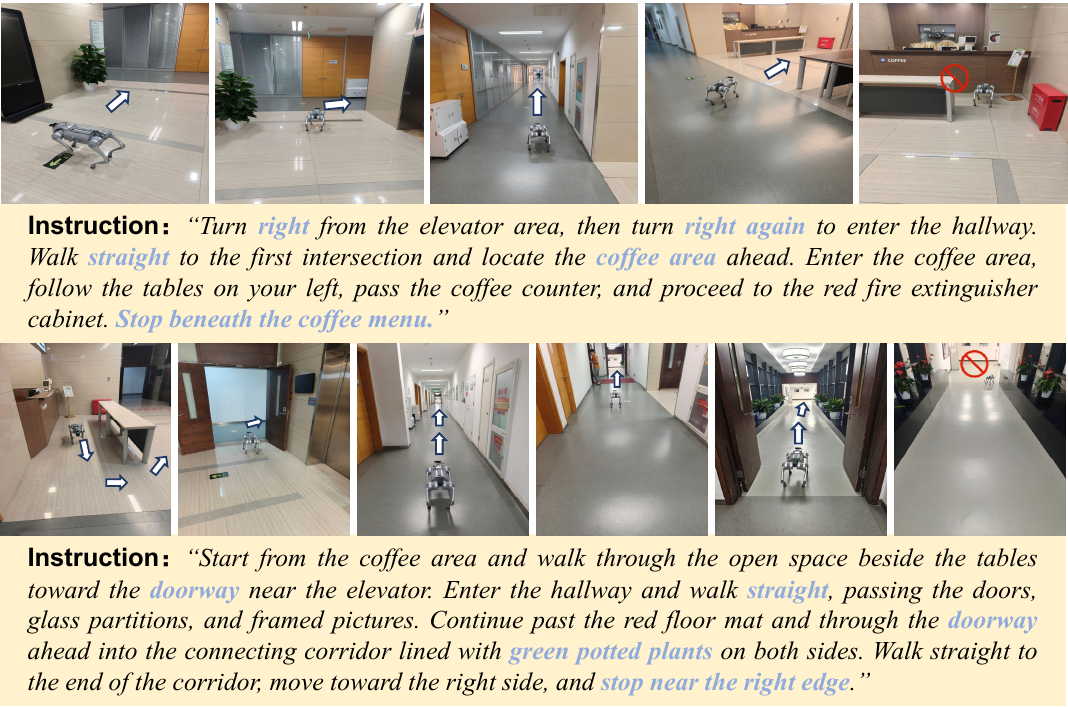}
    \caption{\textbf{Real-world deployment on a Unitree GO2 robot.}}
    \vspace{-1mm}
    \label{fig:realworld}
\end{figure}

\subsection{Real-World Navigation}

We deploy the full policy ($K=4,E=2$) on a Unitree GO2
quadruped robot equipped with an Intel RealSense D435i
RGB-D camera. The policy uses language instructions,
current and historical forward-facing RGB images, and
past actions; depth is used only for low-level collision
avoidance. Figure~\ref{fig:realworld} shows two successful
multi-stage indoor navigation examples, demonstrating
the real-world applicability of our parallel multi-step
decision policy.

\section{Conclusion} \label{sec:conclus}
We presented NavDecision, an efficient multi-step decision model for VLN that combines parallel action prediction with cross-iteration self-referential reinforcement learning. PMAP captures short-horizon action dependencies, while M-SRPO leverages a frozen world model and cross-iteration success memory to internalize long-horizon exploration experience during training without additional deployment-time inference modules. Experiments on R2R-CE and RxR-CE demonstrate competitive navigation performance with a 4B backbone, achieving 64.60\% and 64.57\% success rates, respectively, with up to $22.28\times$ and $18.96\times$ inference speedups.

% Required for CVPR 2027: complete this statement before submission.
% CVPR 2027: required after the main text and before the references.
% This section is excluded from the eight-page main-paper limit.
% Official policy (checked 2026-11-15):
% https://cvpr.thecvf.com/Conferences/2027/AuthorGuidelines
%
% Replace the entire red placeholder below with an accurate, anonymous
% statement. Disclose LLM / generative-AI use in research or writing,
% including editing and grammar. Identify tools, affected parts, and the
% nature and extent of use. Do not invent or omit material uses.
% If a non-public tool's name would reveal your identity, use a generic
% technical description during review, as allowed by the author guidelines.
%
% Use the following sentence ONLY if it accurately describes the work:
% No LLMs or generative AI tools were used in this work.
%
\section*{LLM Usage Statement}

OpenAI ChatGPT (GPT-6 Astra Pro), accessed through the
ChatGPT web application, was used to
assist with English translation and language editing,
terminology refinement, LaTeX formatting and troubleshooting,
and literature search and citation organization.
The tool was used solely as an auxiliary writing and editing
assistant. All research ideas, methodological designs,
experimental procedures, experimental results, and scientific
conclusions were determined by the authors. The authors
reviewed and verified all AI-assisted suggestions and take
full responsibility for the final manuscript.

{
    \small
    
    \bibliographystyle{ieeenat_fullname}
    \bibliography{main}
}


\end{document}